\documentclass[10pt,twocolumn,letterpaper]{article}

\usepackage[pagenumbers]{cvpr} %

\usepackage{comment}
\usepackage{multirow}
\usepackage{multirow}
\usepackage{makecell}
\usepackage{pdflscape}
\usepackage{longtable}

\newcommand{\x}{\mathbf{x}}

\newcommand{\z}{\mathbf{z}}

\usepackage{pifont}%

\newcommand{\itseries}{\fontshape{it}\selectfont} %
\robustify\itseries
\newrobustcmd{\IT}{\itseries}

\graphicspath{ {./Figures/} }

\newcommand{\TSA}[1]{\textcolor{OliveGreen}{#1}}
\newcommand{\TAA}[1]{\textcolor{orange}{#1}}

\definecolor{cvprblue}{rgb}{0.21,0.49,0.74}
\usepackage[pagebackref,breaklinks,colorlinks,allcolors=cvprblue]{hyperref}

\usepackage{amsmath,amssymb,amsfonts}
\usepackage{graphicx}
\usepackage{textcomp}
\usepackage{xcolor}
\usepackage{algorithm}
\usepackage{algorithmic}

\usepackage{url}
\usepackage{algorithm}
\usepackage{algorithmic}
\usepackage{eqparbox}

\def\paperID{13641} %
\def\confName{CVPR}
\def\confYear{2025}

\title{Task-Agnostic Attacks Against Vision Foundation Models}

\author{Brian Pulfer\textsuperscript{1,*} \quad Yury Belousov\textsuperscript{1,*} \quad Vitaliy Kinakh\textsuperscript{1} \quad Teddy Furon\textsuperscript{2} \quad Slava Voloshynovskiy\textsuperscript{1} \\
    \textsuperscript{1}University of Geneva \hspace{3em}
    \textsuperscript{2}University of Rennes, Inria, CNRS, IRISA\\
}

\begin{document}
\maketitle
\def\thefootnote{*}
\footnotetext{These authors contributed equally to this work.}
\def\thefootnote{\arabic{footnote}} %
\def\c{\mathbf{c}}

\begin{abstract}

The study of security in machine learning mainly focuses on downstream task-specific attacks, where the adversarial example is obtained by optimizing a loss function specific to the downstream task.
At the same time, it has become standard practice for machine learning practitioners to adopt publicly available pre-trained vision foundation models, effectively sharing a common backbone architecture across a multitude of applications such as classification, segmentation, depth estimation, retrieval, question-answering and more.
The study of attacks on such foundation models and their impact to multiple downstream tasks remains vastly unexplored.
This work proposes a general framework that forges task-agnostic adversarial examples by maximally disrupting the feature representation obtained with foundation models.
We extensively evaluate the security of the feature representations obtained by popular vision foundation models by measuring the impact of this attack on multiple downstream tasks and its transferability between models.
\end{abstract}

\section{Introduction}

Vision Foundation Models (VFMs) are becoming increasingly popular due to their versatility in handling a wide range of image-based downstream tasks.
Once trained, these models can be minimally fine-tuned to perform
classification and semantic segmentation~\cite{oquab2023dinov2, kirillov2023segment},
object detection~\cite{liu2022swin}, depth estimation~\cite{depthanything, yang2024depth}, 
visual question-answering~\cite{cao2023modularized, tiong2022plug}, image captioning~\cite{mokady2021clipcap}, retrieval~\cite{fernandez2022active}, watermarking~\cite{fernandez2022watermarking} and more.
VFMs are trained with different strategies depending on the Self-Supervised Learning (SSL) framework.
Additionally, some VFMs benefit from the bridge between the text and image modalities as offered by CLIP~\cite{radford2021learning} and follow-up works~\cite{li2022blip,li2022grounded,li2023scaling}.

\noindent VFMs have become a crucial component in many advanced systems due to their versatility, performances, and availability as open-sourced models, which on the other hand raises security concerns. Given their ubiquity, it is crucial to assess their robustness and their security against attacks.

{

\newcommand{\IMGSIZE}{0.15}

\newcommand{\image}[2]{\includegraphics[width=\IMGSIZE\textwidth,keepaspectratio]{Figures/examples/#1/#2.png}}

\newcommand{\cleanimg}[1]{\image{clean}{#1_clean_img}}
\newcommand{\gtimg}[1]{\image{clean}{#1_gt_label}}
\newcommand{\predimg}[1]{\image{clean}{#1_pred_clean}}
\newcommand{\attimg}[2]{\image{#2}{#1_pred_adv}}
\newcommand{\advimg}[2]{\image{#2}{#1_adv_img}}

\begin{figure}[b]
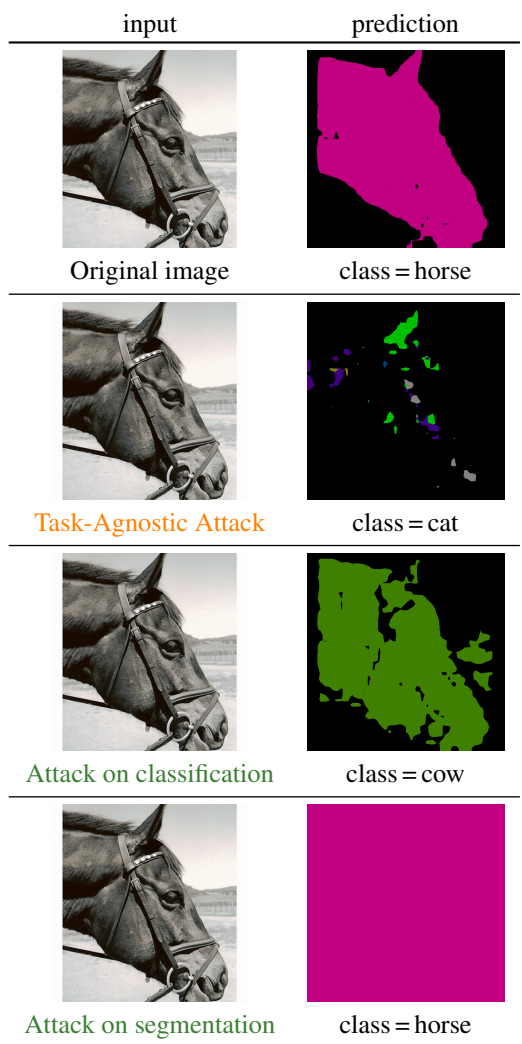

	\centering
	\begin{tabular}{cc}
        input & prediction \\
        \toprule
        \cleanimg{27} & \predimg{27} \\
        Original image & class\,=\,horse \\
        \midrule

        \advimg{27}{patch_tokens} & \attimg{27}{patch_tokens} \\
        \textcolor{orange}{Task-Agnostic Attack} & class\,=\,cat \\
        \midrule

       \advimg{27}{classification} & \attimg{27}{classification} \\
        \textcolor{OliveGreen}{Attack on classification} & class\,=\,cow \\
        \midrule

        \advimg{27}{segmentation} & \attimg{27}{segmentation} \\
        \textcolor{OliveGreen}{Attack on segmentation} & class\,=\,horse \\
        
    \bottomrule
	\end{tabular}
\caption{Adversarial example attacks on DinoV2 ViT-S model. The Task-Agnostic Attack deludes both the segmentation and the classification, contrary to attacks specific to a downstream task.\label{fig:pred_attack}}
\end{figure}
}

This paper defines the {\bf robustness} of VFMs as the system’s ability to produce reliable and consistent outputs for a given downstream task when subjected to standard non-adversarial degradations, such as lossy compression, rotation, flipping, contrast changes, blurring, etc.
The {\bf security} of VFMs, on the other hand, is defined as the system’s resilience in maintaining output integrity for the downstream task when faced with adversarial attacks designed to alter their output. 
Despite the popularity of foundation models, the security aspect remains under-explored~\cite{balestriero2023cookbook}.

Traditional adversarial examples are designed with a specific downstream task in mind. We refer to this family of attacks as {\em task-specific attacks} (TSAs). 
These exploit the particularities of the downstream task head to create adversarial examples that disrupt the model's performance on that task.
A TSA can be non-targeted, aiming to alter the task label to any incorrect output, or targeted, where the task label is changed to a specific, desired output.
While effective, these attacks are limited by their reliance on the characteristics of the downstream tasks.

In contrast, this paper introduces {\em task-agnostic attacks} (TAAs) which do not consider the downstream task head, but rather only target the foundation model. TAAs can also be divided into untargeted attacks, which aim to move the latent representation far from the original one, and targeted attacks, which aim to move the latent to a specific predefined target in the latent space.
~\cref{fig:pred_attack} illustrates that while TSA perform best for the designated downstream task, transferability to other tasks is limited. In contrast, TAA generates an adversarial sample that effectively deceive the model across a range of applications.
This paper investigates the efficiency and transferability of untargeted TAAs on foundation models across various downstream tasks and models.
\begin{figure*}[tb]
\vspace{0 cm}
\centering
        \includegraphics[width=0.89\linewidth]{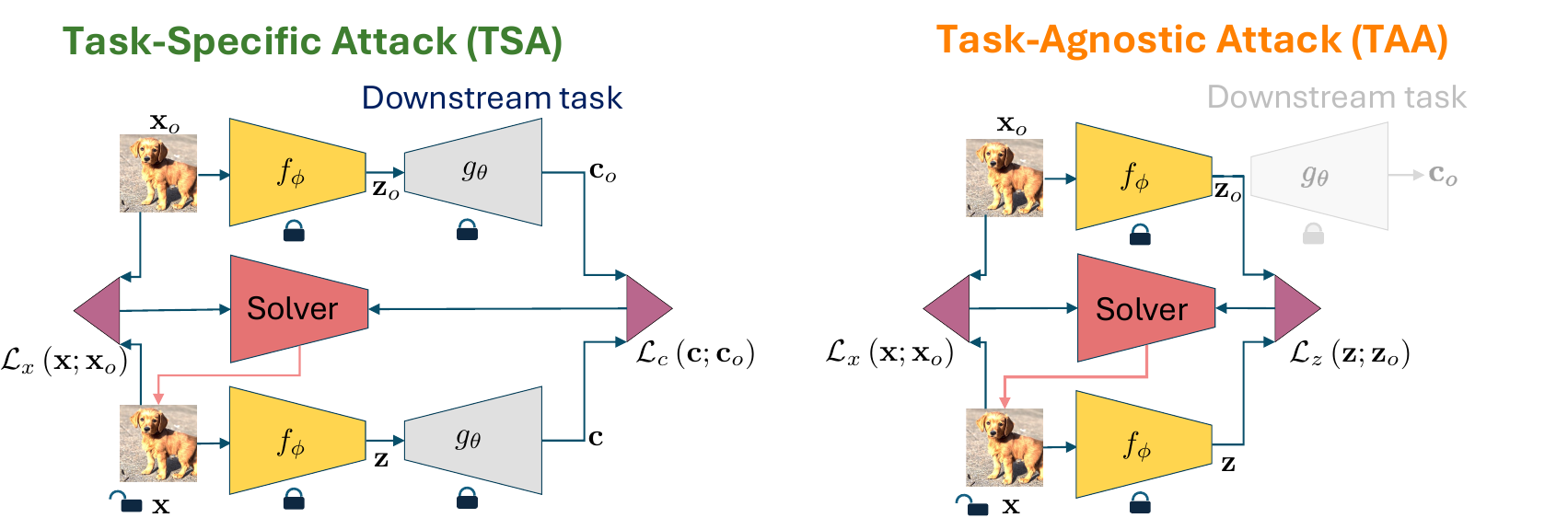}
	\caption{Schematic representation of classic Task-Specific Attack (left) and proposed Task-Agnostic Attack (right). }
	\label{fig:TSA_vs_TAA}	
\end{figure*}

Our first contribution is the design of a new untargeted TAA with variants perturbating different tokens of a Vision Transformer (ViT). %
The second contribution evaluates the pervasion of TAAs in terms of their ability to delude downstream tasks like classification, segmentation, image retrieval, etc.
We also compare TAAs and TSAs for the same image distortion budget.
Finally, we investigate the transferability of the attacks from one white-box source model to a black-box target model.

We demonstrate that task-agnostic attacks generalize effectively across multiple downstream tasks, model sizes and datasets, revealing their universality and their dangerousness against any self-supervised learning and multi-modal system.
These insights emphasize the need for further investigation into the security of foundation models before their widespread deployment in critical applications.

\section{Related Work}

\subsection{Vision Foundation Models}
VFMs provide an unprecedented level of utility and versatility for a wide range of image-based tasks.
VFMs differ in their architecture, their size, and their training set, but the main difference lies in the SSL framework. The training objective is key to create a universal model which feature representations can be applied to a vast amount of downstream tasks.

{\bf MAE}~\cite{He_2022_CVPR} is based solely on the Masked Image Modelling (MIM) objective, where an encoder-decoder architecture learns to reconstruct masked image patches from a few visible patches only.
{\bf MSN}~\cite{10.1007/978-3-031-19821-2_26} combines MIM with Siamese Networks to avoid pixel and token-level reconstructions.
A teacher network computes the representation of a view of an image, while a student network computes the representation of another partially masked view.
MSN is optimized by learning a student that can match the output of the teacher network.
{\bf CAE}~\cite{chen2022context} also learns to predict masked visual tokens from visible tokens by aligning their representations.
{\bf I-JEPA}~\cite{Assran_2023_CVPR} predicts representations of various blocks of an image given only a context block obtained through a specific masking strategy.

{\bf DiNO}~\cite{Caron_2021_ICCV} first proposed self-distillation with no labels.
The outputs of a teacher and student networks are passed through a softmax, and the objective is for the student to minimize the cross-entropy loss between the two probability distributions.
{\bf iBOT}~\cite{zhou2022ibot} builds on top of DiNO, combining the self-distillation strategy with the MIM objective of BEiT~\cite{bao2022beit} within self-distillation.
{\bf DiNOv2}~\cite{oquab2023dinov2} improves on DiNO by using a larger and curated dataset, namely LVD-142M.
It relies on an efficient implementation for training at scale and an advanced SSL framework.
In particular, the authors combine the DiNO cross-entropy loss with the MIM objective used in iBOT.
DiNOv2 also benefits from the Sinkhorn-Knopp batch normalization of SwAV~\cite{varamesh2020selfsupervised}.
The largest model, ViT-G, is distilled into smaller models.

    \begin{table}[t!]
    \caption{List of the VFMs considered in this work. The table reports the SSL framework, the architecture, the number of parameters, and the dimensionality of the feature space.}
    \label{tab:ssl_models}
    \centering
    \begin{tabular}{cccc}
         \toprule
         \textbf{Framework} & \textbf{Architecture} & \textbf{\# Pars. (M)} & \textbf{Dim.}\\
         \midrule
         DiNOv2 &  ViT-S/14 & \multirow{3}{*}{21} & \multirow{3}{*}{384}\\
         \cmidrule(r){2-2}
         DiNO & \multirow{2}{*}{ViT-S/16} & & \\
         MSN & & & \\
         \hline
         CAE & \multirow{2}{*}{ViT-B/14} & \multirow{6}{*}{85} & \multirow{6}{*}{768} \\
         DiNOv2 & & & \\
         \cmidrule(r){2-2}
         DiNO & \multirow{4}{*}{ViT-B/16} & & \\
         iBOT & & & \\
         MAE & & & \\
         MSN & & & \\
         \hline
         CAE & \multirow{2}{*}{ViT-L/14} & \multirow{5}{*}{303} & \multirow{5}{*}{1'024} \\
         DiNOv2 & & & \\
         \cmidrule(r){2-2}
         iBOT & \multirow{3}{*}{ViT-L/16} & & \\
         MAE & & & \\
         MSN & & & \\
         \hline
         I-JEPA & \multirow{2}{*}{ViT-H/14} & \multirow{2}{*}{630} & \multirow{2}{*}{1'280} \\
         MAE & & &\\
         \hline
         DiNOv2 & ViT-G/14 & 1'136 & 1'536 \\
         I-JEPA & ViT-G/16 & 1'011 & 1'408\\
         \bottomrule
    \end{tabular}
\end{table}

Our study considers a set of 18 popular and recent pre-trained VFMs listed in~\cref{tab:ssl_models}.
All models have been pre-trained on the ImageNet dataset~\cite{5206848}, except for DiNOv2, which was pre-trained on LVD-142M~\cite{oquab2023dinov2}, a superset of ImageNet.
We only consider ViTs with patch sizes 14 or 16, since most available VFMs use such a patching strategy.

\subsection{Adversarial Example Attacks}
The seminal works~\cite{DBLP:journals/corr/SzegedyZSBEGF13,DBLP:journals/corr/GoodfellowSS14} on adversarial examples reveal the paradox that Deep Learning classifiers are robust but not secure.
They are robust because their prediction likely remains unchanged when the input image is corrupted by noise addition or JPEG compression.
They are not secure because an attacker can craft a low-amplitude perturbation that deludes the classifier. 
These attacks are typically categorized based on the attacker's knowledge and access to the target model: white-box attacks and black-box attacks.

In the white-box scenario, the attacker has the complete knowledge of the model's architecture and parameters, facilitating the generation of adversarial examples.
They first define a loss function that may combine the classification loss of the input w.r.t. the ground truth class and another most likely class together with the distortion w.r.t. the original image. The attack relies on the gradient of this loss w.r.t. the input computed by backpropagation through the model. 
The typical representatives of white-box attacks include PGD~\cite{DBLP:conf/iclr/MadryMSTV18}, targeting a high Attack Success Rate (ASR) within a given distortion budget,
and DeepFool~\cite{moosavi2016deepfool} or CW~\cite{7958570} minimizing the distortion of a successful adversarial example.
Recent efforts in this area aim at improving the speed of the attack~\cite{DBLP:conf/nips/PintorRBB21,bonnet:hal-03467692}.
The black-box scenario limits the attacker's access to the output of the model, without knowledge of its internals.
This category includes methods such as RayS~\cite{chen2020rays}, SurFree~\cite{maho2021surfree}, and CGBA~\cite{DBLP:conf/iccv/RezaR0D23}.

Adversarial examples delude not only classifiers, but also semantic segmentation and object detection~\cite{xie2017adversarial}, depth estimation~\cite{9207958,Zheng_2024_CVPR}, visual question answering~\cite{sharma2018attend}, captioning~\cite{xu2019exact}, or retrieval~\cite{Tolias_2019_ICCV} models. 
Note that each work proposes an attack specific to the targeted application.
The core process is always the gradient computation but the definition of the adversarial loss is driven by the application.

Since a vast majority of foundation models are open-source, we investigate white-box scenarios with higher emphasis than black-box. At the same time, unlike traditional attacks, our method aims to be universal across applications.

\subsection{Attacks against Vision Foundation Models}
It was shown that complementing classifiers with self-supervision during training improves robustness~\cite{NEURIPS2019_a2b15837}. 
Another study~\cite{chhipa2023selfsupervised} confirms that many VFMs (BarlowTwins, BYOL, SimCLR, SimSiam, SwAV, and DiNO) inherit from this fact a greater robustness against image corruptions.
Robustness is here understood in the statistical point of view, i.e., how the performance smoothly degrades as the distribution shift increases.
This does not encompass adversarial examples crafted with the sole goal of damaging downstream performance.

On the contrary, DINO self-supervision does not improve the security against adversarial example attacks compared to traditional supervision in classification tasks~\cite{rando2022exploring}.
To patch this vulnerability, some works propose to combine self-supervision and adversarial training~\cite{NEURIPS2019_a2b15837,kim2020adversarial}. 

All these works do not study the intrinsic vulnerability of the VFM per se, but once used in a classification task.
One exception is the report~\cite{inkawhich2023adversarial} which designs an attack such that adversarial examples are classified as Out-Of-Distribution samples.
Therefore, whatever the downstream task, the system refuses to process them. 

Our paper studies the security of VFM without assuming any downstream task.

\section{Definition of adversarial attacks}
This section pinpoints the differences between TSAs and TAAs as depicted in ~\cref{fig:TSA_vs_TAA}.

\subsection{Classical Task-Specific Attacks}
Let us consider a model composed by a VFM  $f_{\phi}$ with a downstream task head $g_{\theta}$.
For a given input $\x_o$, let $\c_o = g_{\theta}(f_{\phi}(\x_o))$.
The attacker defines two losses: $\mathcal{L}_x\left(\x; \x_o\right)$ measuring the perceptual distance between an input $\x$ and $\x_o$, and $\mathcal{L}_{c}(\c;\c_o)$
gauging how the task performed on $\x$ differs from the result $\c_o$.
For instance, for the classification task, the result $\c_o$ is a predicted class and $\mathcal{L}_{c}(\c;\c_o) = p(\c_o|\x) - \max_{\c\neq \c_o} p(\c|\x)$. %
Under a distortion budget constraint, the attack looks for the minimizer $\x_a$ of $\mathcal{L}_{c}(\c;\c_o)$ over $\{\x:\mathcal{L}_x\left(\x; \x_o\right)\leq\epsilon\}$, with $\c = g_{\theta}(f_{\phi}(\x))$.
An alternative is to solve the dual problem, i.e. the attack looks for the minimizer $\x_a$ of $\mathcal{L}_x\left(\x; \x_o\right)$ under the constraint $\mathcal{L}_{c}(\c;\c_o)\leq \tau_c$, with $\c = g_{\theta}(f_{\phi}(\x))$.

The adversarial example $\x_a$ crucially depends on the downstream task via functions $g_{\theta}$ and $\mathcal{L}_{c}$.
This implies that the attacker knows the downstream task head (white-box) and that $\x_a$ may not be adversarial for different tasks.

\subsection{Proposed Task-Agnostic Attack}

In contrast to TSAs, we target a new type of task-agnostic attack when the downstream head or even the downstream task are unknown.
The objective is to forge \emph{pervasive} adversarial examples, i.e. jeopardizing a large spectrum of downstream tasks.
The proposed TAA aims at maximally perturbing the features obtained from the VFM backbone.

Given a foundation model \( f_\phi: \mathcal{X} \rightarrow \mathcal{Z} \), the features of VFM are computed as $\mathbf{z} = f_{\phi}(\x) \in \mathcal{Z}$. For a ViT, $\mathbf{z}$ may represent any aggregating function of the output class and patch tokens.
We compute latent embeddings \( \mathbf{z}_j = f_\phi(\mathbf{x}_j) \) for some training samples \( \{\mathbf{x}_j\}_{j=1}^{N_T} \), which we use to compute an empirical mean ${\boldsymbol \mu} = \frac{1}{N_T} \sum_{k=1}^{N_T} \mathbf{z}_k$ which can be used to center features extracted with the VFM as $\tilde{\mathbf{z}} = \mathbf{z} - {\boldsymbol \mu}$.

For a given input $\x_o$, we denote $\mathbf{z}_o = f_{\phi}(\x_o) \in \mathcal{Z}$. The loss function is now defined in the feature space $\mathcal{Z}$ as:
\begin{equation}
    \label{eq:3}
     \mathcal{L}_z\left(\z; \z_o\right) = \text{cos\_sim}({\tilde{\bf z}},{\tilde{\bf z}}_o).
\end{equation}

We notice that features extracted from some foundation models are not centered around the origin, making the mean-centering process necessary to compute meaningful cosine similarities. We report results without this pre-processing step in supplementary material.

Again, two alternatives are 1) minimize $\mathcal{L}_z\left(\z; \z_o\right)$ under the distortion budget constraint $\mathcal{L}_x\left(\x; \x_o\right)\leq\epsilon$, or 2) minimize the distortion $\mathcal{L}_x\left(\x; \x_o\right)$ under an objective constraint $\mathcal{L}_z\left(\z; \z_o\right)<\tau_z$.
We prefer the first option as it eases the fair comparison of the attacks by monitoring the Attack Success Rate for a given distortion budget $\epsilon$. Under a distortion budget constraint, the attack looks for the minimizer $\x_a$ of $\mathcal{L}_x\left(\x; \x_o\right)$ under the constraint $\mathcal{L}_z\left(\z; \z_o\right)<\tau_z$. Algorithm~\ref{alg:taa} summarizes our method.

\begin{algorithm}
\caption{Task-Agnostic Attack}
\label{alg:taa}
\begin{algorithmic}[1]
\STATE \textbf{Input}: $\mathbf{x}_o$: original image, \( \{\mathbf{x}_j\}_{j=1}^{N_T} \): training images, $f_\phi$: backbone model, $\epsilon_\infty$: maximum per-pixel perturbation, $\alpha$: step size, $N$ number of steps;

\COMMENT{ // Compute mean vector from set of images for centering. $sg$: stop gradient}

\STATE $\{\mathbf{z}_j\}_{j=1}^{N_T}  \leftarrow \{sg(f_\phi( \mathbf{x}_j))\}_{j=1}^{N_T}$

\STATE ${\boldsymbol \mu} \leftarrow \frac{1}{N_T} \sum_{k=1}^{N_T} \mathbf{z}_k$

\COMMENT{ // initialize attack and target}

\STATE $\mathbf{x}_a \leftarrow \mathbf{x}_o$
\STATE $\tilde{\mathbf{z}}_o \leftarrow \text{sg}(f_\phi(\mathbf{x}_o)) - \boldsymbol \mu $ 

\COMMENT{ // Run PGD in feature space using cos-sim}

\FOR{$t = 0, \ldots, N-1$}
    \STATE $\tilde{\mathbf{z}} \leftarrow 
  f_\phi(\mathbf{x}_a) - \boldsymbol \mu$
    \STATE $\mathcal{L}_z \leftarrow \text{cos\_sim}(\tilde{\mathbf{z}}, \tilde{\mathbf{z}}_o)$ 
    \STATE $\mathbf{x}_a \leftarrow \mathbf{x}_a - \alpha \times \nabla_{\x_a}\mathcal{L}_z
    $ 
    \STATE $\mathbf{x}_a \stackrel{\text{constraints}}{\longleftarrow} \mathbf{x}_a$ \COMMENT{// impose constraints via ${\epsilon}_{\infty}$}
\ENDFOR
\STATE \textbf{Return}: Attacked image $\mathbf{x}_a$\\
\end{algorithmic}
\end{algorithm}

\paragraph{Implementation}
We use Projected Gradient Descent (PGD)~\cite{DBLP:conf/iclr/MadryMSTV18} as our solver. We set the step size $\alpha = 0.0004$, the total number of optimization steps to $50$, and $\epsilon_\infty = \frac{8}{255}$. %
The adversarial tensors are then clipped and quantized to get real adversarial images, which are stored in \textit{png} format and re-loaded for evaluation. This post-processing can result in slightly higher or lower $\mathcal{L}_x\left(\x; \x_o\right)$ with respect to the target $\tau_z$.
To carry out TAAs, we developed a highly customizable PyTorch~\cite{Paszke_PyTorch_An_Imperative_2019} code which comes with public pre-trained SSL models used in this work. Source code is available at \url{https://github.com/BrianPulfer/fsaa}.

\section{Experimental setup}
\label{sec:exp}

\paragraph{Models} ~\cref{tab:ssl_models} lists the models of the experimental setup. For each model, we use its default normalization to convert images to tensors.

\paragraph{Datasets and metrics for downstream tasks}
We use popular datasets and metrics for each downstream task:
\begin{itemize}
    \item Classification: Accuracy on PascalVOC~\cite{pascal-voc-2012} or ImageNette~\cite{Howard_Imagenette_2019} 
    \item Segmentation: Mean Intersection over Union (mIoU) on PascalVOC~\cite{pascal-voc-2012}
    \item Visual Question Answering: Accuracy on VQAv2~\cite{balanced_vqa_v2}
    \item Image Captioning: BLEU-4, METEOR, ROUGE-L and CIDEr scores on COCOCaptions~\cite{chen2015microsoft}
    \item Image Retrieval: Mean Average Precision (mAP) on the Revisited Oxford buildings dataset \cite{RITAC18}.
\end{itemize}

\paragraph{Metrics for attacks\\}

\textbf{Absolute efficiency}: We measure the drop in performance with the metrics and datasets mentioned above.

\textbf{Relative efficiency}: We define the \emph{relative efficiency} $\eta$ of a TAA w.r.t. a downstream task as a percentage, where $0\% $ indicates no impact on task accuracy, and $100\% $ reflects an impact equivalent to the respective TSA. Formally,
\begin{equation}
    \eta = 100\times\frac{\text{perf(TAA)} - \text{perf(No\,attack)}}{\text{perf(TSA)} - \text{perf(No\,attack)}},
    \label{eq:RelEff}
\end{equation}
where $\text{perf}$ is a performance metric of the downstream task. 

\textbf{Image quality}: For a fair comparison of the attacks, we set a target Peak Signal-to-Noise Ratio $\text{PSNR} = 10\log_{10} \frac{255^2}{\text{MSE}}$ with $\text{MSE} =\frac{1}{3HW}\|\x_o - \x_a \|^2$, where $\x\in[0,1]^{H \times W\times 3}$. In experiments we set ${\text{PSNR}}=40$ dB.

\section{Experimental results on VFMs}

\subsection{Robustness \textit{vs.} Security}

Foundation models enable the \emph{generalisation} of the downstream heads for many applications.
Systems built on foundation models demonstrate remarkable performance across various downstream tasks under clean conditions.
They also enable great \emph{robustness} against common editing transformation such as flipping, blurring or JPEG compression.
The accuracy smoothly degrades as the shift between distributions of the training  and testing data gets larger.
This motivates the analysis of the security level which would have been useless if not robust first and foremost. 
Downstream heads are learned on the PascalVOC training dataset for \emph{classification} and \emph{segmentation} each while the model backbone DiNOv2 ViT-S is kept frozen.
~\cref{tab:robustness} gauges the robustness against common image processing and the security against Task-Specific Attacks over images from the validation dataset and $\text{PSNR}=40$ dB.
For these two applications, the global system is clearly robust yet not secure against a TSA.

\begin{table}[htbp]
\caption{Classification and  segmentation performance for a set of distortions (\textit{robustness}) and TSA (\textit{security}).}
\label{tab:robustness}
\centering
\begin{tabular}{cccc}
\toprule
\textbf{Transform}  & \textbf{Classification} & \textbf{Segmentation} \\
\midrule

\IT clean performance & \IT 96.3  &  \IT 81.4 \\

\midrule

horizontal flip & 95.9 & 81.3 \\  %
vertical flip & 87.9 & 61.0 \\ %
wiener filter $\text{size}=21$ & 95.7 & 76.6\\
blur $\text{kernel\_size}=21$ & 94.5 & 80.3 \\  %
jpeg $\text{quality}=50$ & 95.8 & 81.2 \\
grayscale & 96.3 & 81.3 \\
rotation by 90\textdegree & 88.5 & 65.7 \\
resize to $98 \times 98$ & 88.5 & 61.3 \\ 
brightness $\text{factor}=2$ & 95.3  & 81.2 \\
contrast $\text{factor}=2$ & 95.8 & 80.9 \\
hue $\text{factor}=0.5$ & 95.9 & 81.1 \\

\midrule
Task-Specific Attack & 0.0 & 11.8 \\
\bottomrule
\end{tabular}
\end{table}

\subsection{TAA \textit{vs.} TSA}
\label{sec:TAAvsTSA}
\begin{table}[bp]
\caption{mAP on the image retrieval with DiNOv2 on the R-Oxford dataset. PSNR is set to 40 dB.}
\label{tab:ir}
\centering
\begin{tabular}{cccc}
\toprule
\textbf{Model} & \textbf{Difficulty} & \textbf{Clean$\uparrow$} & \textbf{TAA$\downarrow$} \\
\midrule
\multirow{3}{*}{DiNOv2 ViT-S} & Easy & 82.1 & \textcolor{orange}{0.6} \\
    & Medium & 67.3 & \textcolor{orange}{1.1} \\
    & Hard & 41.4 & \textcolor{orange}{0.5} \\
\hline
\multirow{3}{*}{DiNOv2 ViT-B} & Easy & 85.7 & \textcolor{orange}{1.6} \\
    & Medium & 71.7 & \textcolor{orange}{2.0} \\
    & Hard & 49.4 & \textcolor{orange}{0.8} \\
\hline
\multirow{3}{*}{DiNOv2 ViT-L} & Easy & 87.9 & \textcolor{orange}{1.0} \\
    & Medium & 74.4 & \textcolor{orange}{1.9} \\
    & Hard & 53.1 & \textcolor{orange}{1.3} \\
\hline
\multirow{3}{*}{DiNOv2 ViT-G} & Easy & 85.2 & \textcolor{orange}{7.4} \\
    & Medium & 72.9 & \textcolor{orange}{7.5} \\
    & Hard & 52.3 & \textcolor{orange}{4.2} \\
\bottomrule
\end{tabular}
\end{table}

\begin{table*}[t]
\caption{Absolute and relative efficiency~\eqref{eq:RelEff} of attacks against classification and segmentation by two DiNOv2 models on PascalVOC validation set. TAAs are more pervasive across tasks while TSAs are more harmful w.r.t. the targeted task. Target PSNR = 40 dB. The arrow indicates the direction of success for attacker.}
\label{tab:downstream_attack}
\centering
\begin{tabular}{ccccc}
\toprule
\textbf{Backbone} & \textbf{Attack}  &  \textbf{Type}      & \textbf{Classification abs$\downarrow$ (rel$\uparrow$)} & \textbf{Segmentation abs$\downarrow$ (rel$\uparrow$)} \\
\midrule
\multirow{6}{*}{ViT-S} & \IT No attack & & \IT 96.3 (0\%) & \IT 81.4 (0\%) \\
&\textcolor{orange}{Class token} & \textcolor{orange}{TAA} & \textcolor{orange}{7.9 (92\%)}& \textcolor{orange}{19.2 (86\%)} \\
&\textcolor{orange}{Patch tokens} & \textcolor{orange}{TAA} & \textcolor{orange}{\bf 0.1 (100\%)}& \textcolor{orange}{\bf 11.6 (97\%)} \\
&\textcolor{orange}{Class+patch tokens} & \textcolor{orange}{TAA} & \textcolor{orange}{2.0 (98\%)} & \textcolor{orange}{13.3 (94\%)} \\
&\textcolor{OliveGreen}{Classification}& \textcolor{OliveGreen}{TSA} & \textcolor{OliveGreen}{\bf 0.0 (100\%)} & \textcolor{OliveGreen}{19.8 (85\%)} \\
&\textcolor{OliveGreen}{Segmentation} & \textcolor{OliveGreen}{TSA} & \textcolor{OliveGreen}{40.6 (58\%)} & \textcolor{OliveGreen}{\bf 9.2 (100\%)} \\
\midrule
\multirow{6}{*}{ViT-B} &\IT No attack & & \IT 97.0 (0\%) & \IT 80.8 (0\%) \\
&\textcolor{orange}{Class token} & \textcolor{orange}{TAA} & \textcolor{orange}{11.8 (88\%)}& \textcolor{orange}{23.5 (80\%)}\\
&\textcolor{orange}{Patch tokens} & \textcolor{orange}{TAA} & \textcolor{orange}{\bf 0.0 (100\%)}& \textcolor{orange}{\bf 7.5 (102\%)}\\
&\textcolor{orange}{Class + patch tokens} & \textcolor{orange}{TAA} & \textcolor{orange}{2.1 (98\%)}& \textcolor{orange}{11.8 (96\%)}\\
&\textcolor{OliveGreen}{Classification} & \textcolor{OliveGreen}{TSA} & \textcolor{OliveGreen}{\bf 0.0 (100\%)}& \textcolor{OliveGreen}{14.1 (93\%)}\\
&\textcolor{OliveGreen}{Segmentation} & \textcolor{OliveGreen}{TSA} & \textcolor{OliveGreen}{43.9 (55\%)}& \textcolor{OliveGreen}{\bf 8.9 (100\%)}\\
\bottomrule
\end{tabular}
\end{table*}

With the identical setup as previously described, ~\cref{tab:downstream_attack} provides a comparison of TAAs and TSAs. The conclusions are clear: TSAs present a more effective attack approach compared to TAAs when examining the performance impacts on the targeted applications. Yet, TSAs exhibit limited pervasiveness.
For example, the TSA directed at segmentation reduces the accuracy of the classification task only by around half.
In contrast, TAAs negatively impact both tasks equally.

~\cref{tab:downstream_attack} shows that TAAs can reach a minimum relative efficiency of $80\% $. Notably, TAAs that compromise patch tokens demonstrate near equivalent efficiency as TSAs.

\subsection{Extending TAAs to other downstream tasks}

\paragraph{Image retrieval:} We measure the efficacy of TAAs for the image retrieval task on the revisited Oxford dataset~\cite{RITAC18} using DiNOv2 backbones and a target PSNR of 40 dB. ~\cref{tab:ir} reports the mean average precision. Interestingly, image retrieval capabilities do not improve beyond a ViT-L model, yet, the adversarial robustness to TAAs does improve with a larger ViT-G architecture. Still, all model performances are severly degraded.

\begin{figure}[htbp]
    \centering
    \begin{subfigure}[b]{\linewidth}
        \centering
        \includegraphics[width=0.97\linewidth]{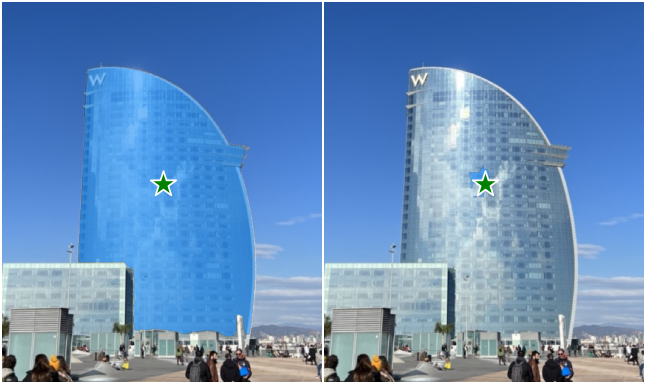}
    \end{subfigure}
    \begin{subfigure}[b]{\linewidth}
        \centering
        \includegraphics[width=0.97\linewidth]{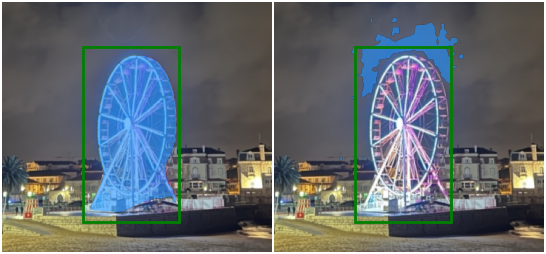}
    \end{subfigure}
    \caption{Our TAA deludes Segment-Anything-Model~\cite{kirillov2023segment}. Original (left) and adversarial (right, PSNR = 40 dB) images.}
    \label{fig:att_sam}
\end{figure}

\begin{table}[htbp]
\caption{Accuracy in zero-shot classification on ImagenetV2 for CLIP model with PSNR = 40 dB.}
\label{tab:zero_shot_accuracy_clip}
\centering
\begin{tabular}{lcc}
\toprule
& \textbf{No attack $\uparrow$} & \textcolor{orange}{TAA $\downarrow$} \\
\midrule
Top-1 & \IT 55.95 &\textcolor{orange}{0.03} \\
Top-5 & \IT 83.39 & \textcolor{orange}{0.21} \\
\bottomrule
\end{tabular}
\end{table}

\paragraph{Zero-shot segmentation: } In ~\cref{fig:att_sam} the TAA targets the patch tokens of the ViT-H encoder in the Segment Anything Model (SAM)~\cite{kirillov2023segment}, without relying on the prompt encoder and mask decoder.
It shows segmentation masks before and after the attack with the same query (the green star point or the bounding box).

\paragraph{Zero-shot classification:} 
Recent Vision-Language Models (VLM) \cite{beyer2024paligemmaversatile3bvlm, xiao2311florence, liu2024visual, wan2024locca, li2022blip, li2023blip, xue2024xgen} combine a pre-trained VFM as an image encoder together with a pre-trained language model, the latter usually involving more parameters and being computationally more demanding than the former.
TAAs on the VFM compromise performances of VLMs without requiring computationally expensive evaluation of the language model.

~\cref{tab:zero_shot_accuracy_clip} reports how vulnerable is \emph{zero-shot classification} based on CLIP~\cite{radford2021learningclip}.
Our results are consistent with previous TSAs designed in~\cite{mao2022clipunderstanding, Fort2021CLIPadversarial}.
However, we show that this lack of security holds true for TAA as well.

\paragraph{Image captioning and VQA:} We run TAAs against PaliGemma~\cite{beyer2024paligemmaversatile3bvlm} to measure the drop in performance for \emph{image captioning} over the COCO validation set~\cite{chen2015microsoft} and \emph{visual question answering} over the VQAv2 dataset~\cite{VQA, balanced_vqa_v2}. In both cases, we simply aim at maximally perturbing the feature representation obtained with SigLip, and do not make use of the much larger Gemma~\cite{gemmateam2024gemmaopenmodelsbased, gemmateam2024gemma2improvingopen} language model.

\begin{figure*}[htbp]
    \centering
\includegraphics[width=0.96\linewidth]{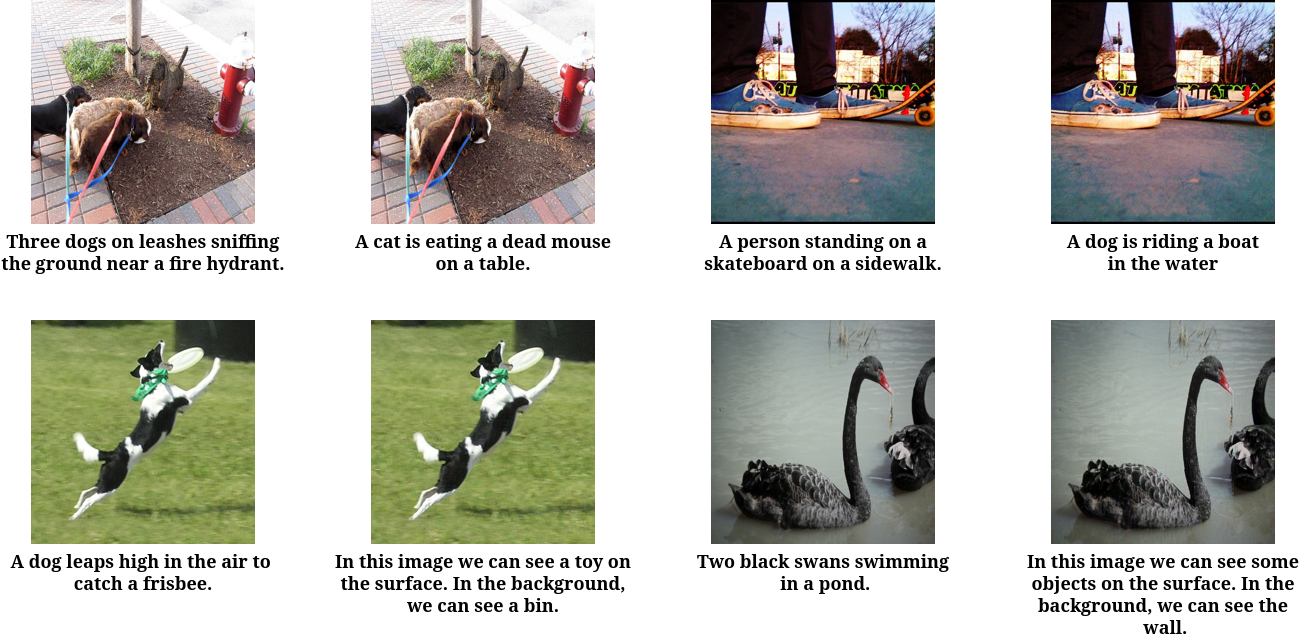}
    \caption{Examples of regular (left) and adversarial (right) captions obtained with TAAs attacking the VFM of PaliGemma. PSNR is 40 dB.}
    \label{fig:vlm-cap-reg}
\end{figure*}

~\cref{tab:vlm-cap} shows quantitative results illustrated with captioning examples in ~\cref{fig:vlm-cap-reg}.
For both downstream tasks, a severe drop in performances occurs even with lower levels of PSNR.
Particularly, we record a higher drop in accuracy for VQAv2 for the less trivial "other" category of questions, and a dramatic drop in CIDEr on the captioning dataset.

\begin{table*}[htbp]
    \centering
    \caption{TAAs against the SigLip VFM of the PaliGemma VLM.  Impact on captioning and question answering. For all metrics, lower values mean more harmful attack. Drop in performance is measured for increasingly stronger perturbations. %
    }
    \begin{tabular}{c|cccc|ccc}
    \toprule
     {\bf Attack} & \multicolumn{4}{c|}{\bf Captioning COCO} & \multicolumn{3}{c}{\bf Question answering VQAv2}\\
         PSNR & BLEU-4 $\downarrow$ &  METEOR $\downarrow$ &  ROUGE-L $\downarrow$ &  CIDEr $\downarrow$ & number $\downarrow$ & yes/no $\downarrow$ & other $\downarrow$\\
         \midrule
         \textit{No attack} & \textit{29.6} & \textit{30.3} & \textit{59.0} & \textit{131.4} & \textit{72.5} & \textit{95.9} & \textit{76.9}\\
         \textcolor{orange}{45 dB} & \textcolor{orange}{5.8} & \textcolor{orange}{17.3} & \textcolor{orange}{32.2} & \textcolor{orange}{33.0} & \textcolor{orange}{50.6} & \textcolor{orange}{83.4} & \textcolor{orange}{53.4}\\
         \textcolor{orange}{40 dB} & \textcolor{orange}{3.8} & \textcolor{orange}{13.6} & \textcolor{orange}{27.2} & \textcolor{orange}{16.4} & \textcolor{orange}{38.7} & \textcolor{orange}{76.0} & \textcolor{orange}{41.6}\\
         \textcolor{orange}{35 dB} & \textcolor{orange}{1.9} & \textcolor{orange}{9.7} & \textcolor{orange}{22.3} & \textcolor{orange}{3.6} & \textcolor{orange}{25.6} & \textcolor{orange}{67.5} & \textcolor{orange}{28.5} \\
         \bottomrule
    \end{tabular}
    \label{tab:vlm-cap}
\end{table*}

\section{Results on transferability between models}
So far, TAAs targeted a given VFM in a white-box scenario.
This section extends the attack surface to the black-box scenario.
In the adversarial example literature, transferability attacks are sometimes efficient to delude an unknown classifier.
Our setup takes the same spirit.
The attacker mounts the TAA against this  white-box \emph{source} model in the hope that this adversarial example also deludes the application based on the  black-box \emph{target} VFM.

\subsection{Transferability across VFM backbones}

\begin{figure*}[ht]
    \centering
    \begin{subfigure}{0.48\textwidth}
        \centering
        \includegraphics[width=\linewidth]{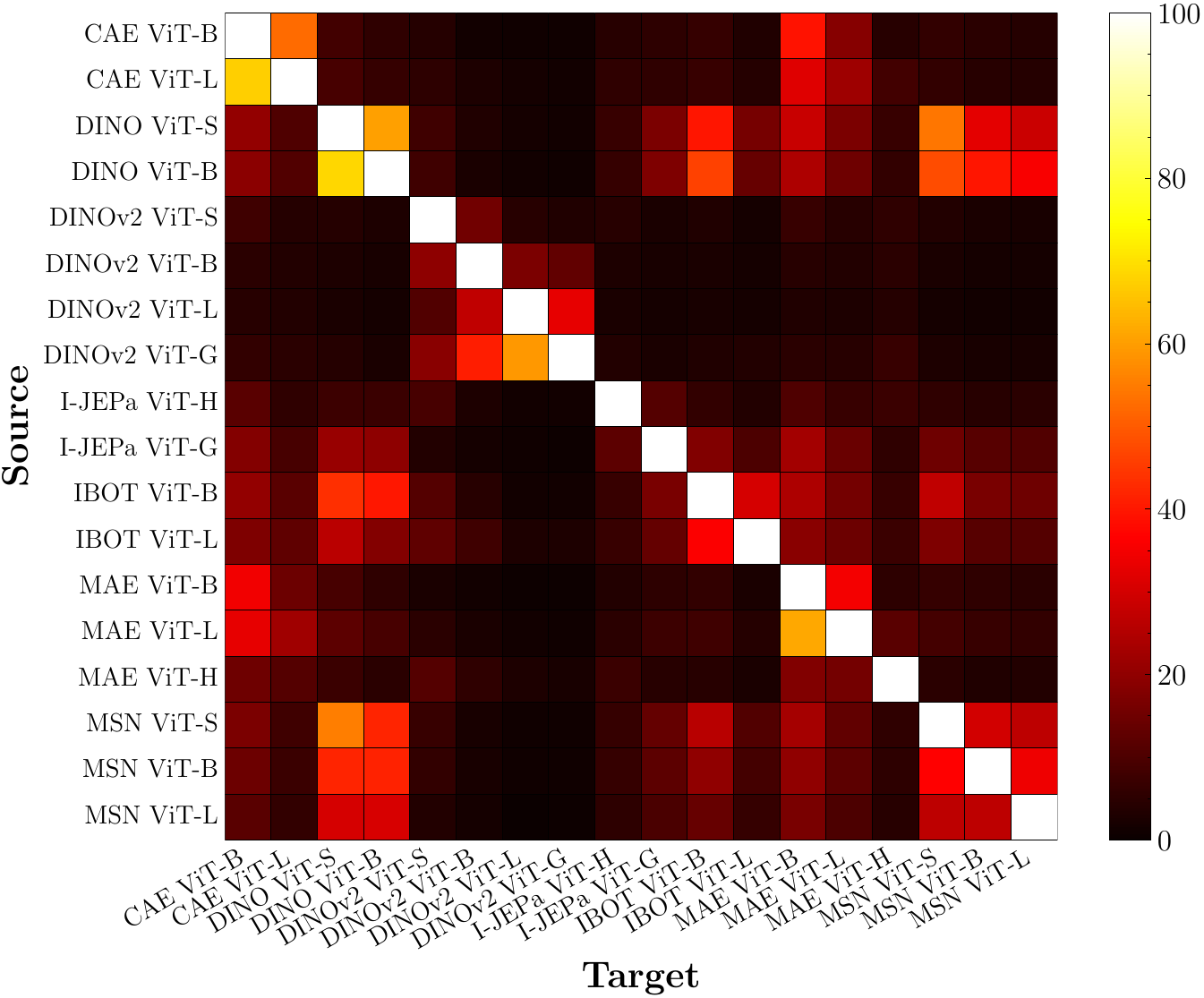}
        \caption{\textcolor{OliveGreen}{Task Specific Attacks (TSA)}}
    \end{subfigure}
    \begin{subfigure}{0.48\textwidth}
        \centering
        \includegraphics[width=\linewidth]{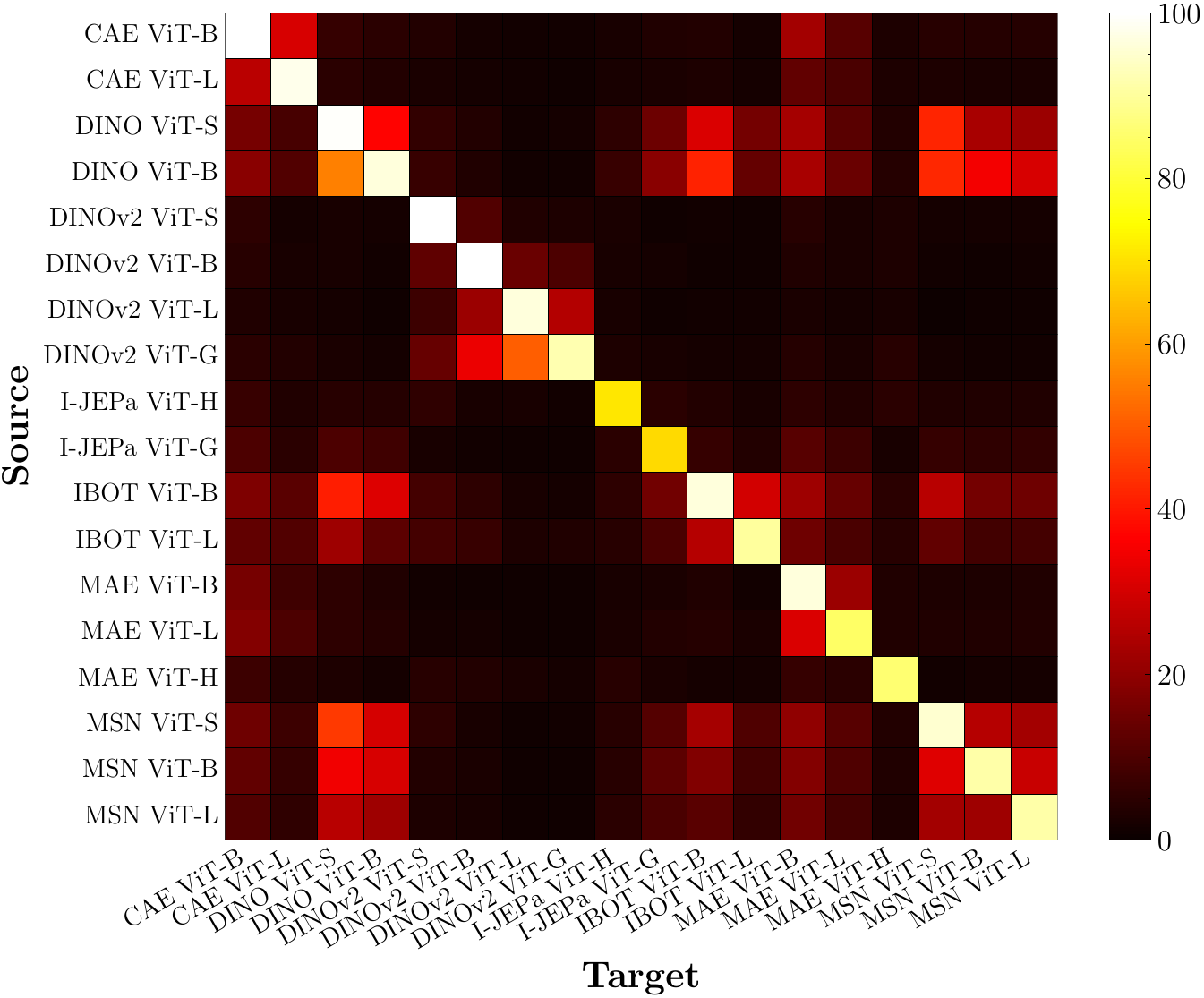}
        \caption{\textcolor{orange}{Task Agnostic Attacks (TAA)}}
    \end{subfigure}
    \caption{Comparison of relative efficiency~\eqref{eq:RelEff} of TAAs (right) with respect to TSAs (left) averaged over classification and semantic segmentation tasks. TAAs perform comparably to TSAs across models and tasks.}
    \label{fig:models-table}
\end{figure*}

For all backbones, a downstream head is trained for classification and semantic segmentation.
For classification, we train linear heads on the ImageNet train set, but we evaluate attacks on the smaller Imagenette dataset~\cite{Howard_Imagenette_2019}.
For semantic segmentation, we concatenate the \texttt{CLS} token to every patch token along the feature dimension and use bi-cubic interpolation followed by a 2-dimensional convolution to obtain predictions for the original image size.
Again, TSAs use the trained heads whereas TAAs do not.

~\cref{fig:models-table} presents the relative efficiency as defined in ~\cref{eq:RelEff} for TAA compared to TSA, averaged across the two proposed tasks. TAA demonstrates near equivalent efficiency to TSA, affirming its effectiveness even without downstream task knowledge. TAAs are more successful against smaller models, and we attribute this to the lower dimensionality of the feature space.

Both TSA and TAA exhibit moderate transferability across different models, with DINO and MSN being the most susceptible to transfer effects. However, this transferability is rare and lacks a clear pattern that would explain why it occurs specifically between these models.

Furthermore, transferability tends to be more likely when both source and target models belong to the same pre-training framework but differ in size. Even within these model families, however, transfer is not guaranteed and remains inconsistent.

\subsection{Transferability to fine-tuned VFMs}
While it is rare for practitioners to pre-train a VFM, it is common practice nowadays to fine-tune an existing open-source VFM for a specific task.
Parameter efficient fine-tuning techniques such as LoRA and variants~\cite{hu2021lora, dettmers2024qlora, lialin2023relora} have become increasingly popular, as they usually represent a more convenient solution compared to full fine-tuning.
Learning more about the degree of transferability of TSAs and TAAs crafted on the publicly available foundation models to their fine-tuned versions is crucial to better understand and design safer machine learning systems.
To this end, we fully fine-tuned and LoRA fine-tuned three VFMs on classification for ImageNet.
LoRA fine-tuning creates query and value matrices adapters in self-attention layers with rank $r=8$, and enforces a dropout of 0.1 during training.

~\cref{tab:finetuning} highlights classification accuracies against TSAs and TAAs using the original VFM as the source model to attack the fine-tuned target model. A LoRA-based fine-tuning is not sufficient to defend against either TSAs or TAAs,
however, the effectiveness of attacks against fully fine-tuned VFMs and downstream heads is greatly reduced.

\begin{table}[htbp]
    \centering
    \caption{Classification accuracy over ImageNette for fine-tuned models and transferable Task-Specific / Task-Agnostic Attacks.}
    \begin{tabular}{ccccc}
        \hline
         \textbf{Attack} & \textbf{Backbone} & \textbf{No FT} & \textbf{LoRA FT} & \textbf{Full FT}  \\
         \hline
         \textcolor{orange}{TAA} & \textcolor{orange}{DiNOv2} & \textcolor{orange}{0.2} & \textcolor{orange}{10.4} & \textcolor{orange}{92.4} \\
         \textcolor{orange}{TAA} & \textcolor{orange}{MAE} & \textcolor{orange}{6.2} & \textcolor{orange}{15.6} & \textcolor{orange}{87.3} \\
         \textcolor{orange}{TAA} & \textcolor{orange}{MSN} & \textcolor{orange}{9.9} & \textcolor{orange}{14.0} & \textcolor{orange}{84.5} \\
         
         \textcolor{OliveGreen}{TSA} & \textcolor{OliveGreen}{DiNOv2} & \textcolor{OliveGreen}{4.4} & \textcolor{OliveGreen}{5.0} & \textcolor{OliveGreen}{92.0} \\
         \textcolor{OliveGreen}{TSA} & \textcolor{OliveGreen}{MAE} & \textcolor{OliveGreen}{5.9} & \textcolor{OliveGreen}{6.1} & \textcolor{OliveGreen}{77.1} \\
         \textcolor{OliveGreen}{TSA} & \textcolor{OliveGreen}{MSN} & \textcolor{OliveGreen}{5.1} & \textcolor{OliveGreen}{6.0} & \textcolor{OliveGreen}{75.7} \\
         \hline
    \end{tabular}
    \label{tab:finetuning}
\end{table}

\section{Conclusion}
This work investigates the robustness and security of the currently most popular pre-trained SSL vision foundation models.
It introduces a family of adversarial example attacks that is task-agnostic and exploits publicly available foundation models to craft adversarial attacks that maximally perturb their feature representation.

Firstly, we find that a TAA performs comparably to its TSA counterpart, while transfering better to other tasks.
Secondly, we show that TAAs are capable of disrupting classification, semantic segmentation, zero-shot classification, image captioning, visual question answering, and image retrieval systems.
Finally, when it comes to transferability across different or fully fine-tuned models, we observe low transferability for both TSAs and TAAs, however, we find that fine-tuning through low rank adapters does not protect from attacks crafted on the non-tuned publicly available model. %

This research aims to enhance AI security by identifying vulnerabilities, promoting defenses that protect models in open-source environments: (i)  {\bf Awareness for Accelerated Defense.} By sharing these findings, we aim to drive proactive research on defenses, reducing potential risks; (ii) {\bf  Limited Risk Scope.} Our attack lacks resilience to transformations (e.g., resizing, flipping), limiting its real-world impact; (iii) {\bf  Potential Defensive Strategies.} Adversarial security can be achieved through adversarial training~\cite{NEURIPS2019_a2b15837,kim2020adversarial,pulfer2024robustness} or could be incorporated ad hoc~\cite{carlini2023free}.

{
    \small
    \bibliographystyle{ieeenat_fullname}
    \bibliography{main}
}

\clearpage
\setcounter{section}{0}
\renewcommand{\thesection}{\Alph{section}}
\maketitlesupplementary

\section{Ablation study}

\subsection{Uncentered features in VFMs}

We observe that features extracted from foundation models are not inherently centered, making cosine similarity loss unsuitable unless appropriate centering is applied.

\begin{figure*}[htbp]
    \centering
    \begin{subfigure}{0.49\textwidth}
        \centering
        \includegraphics[width=\linewidth]{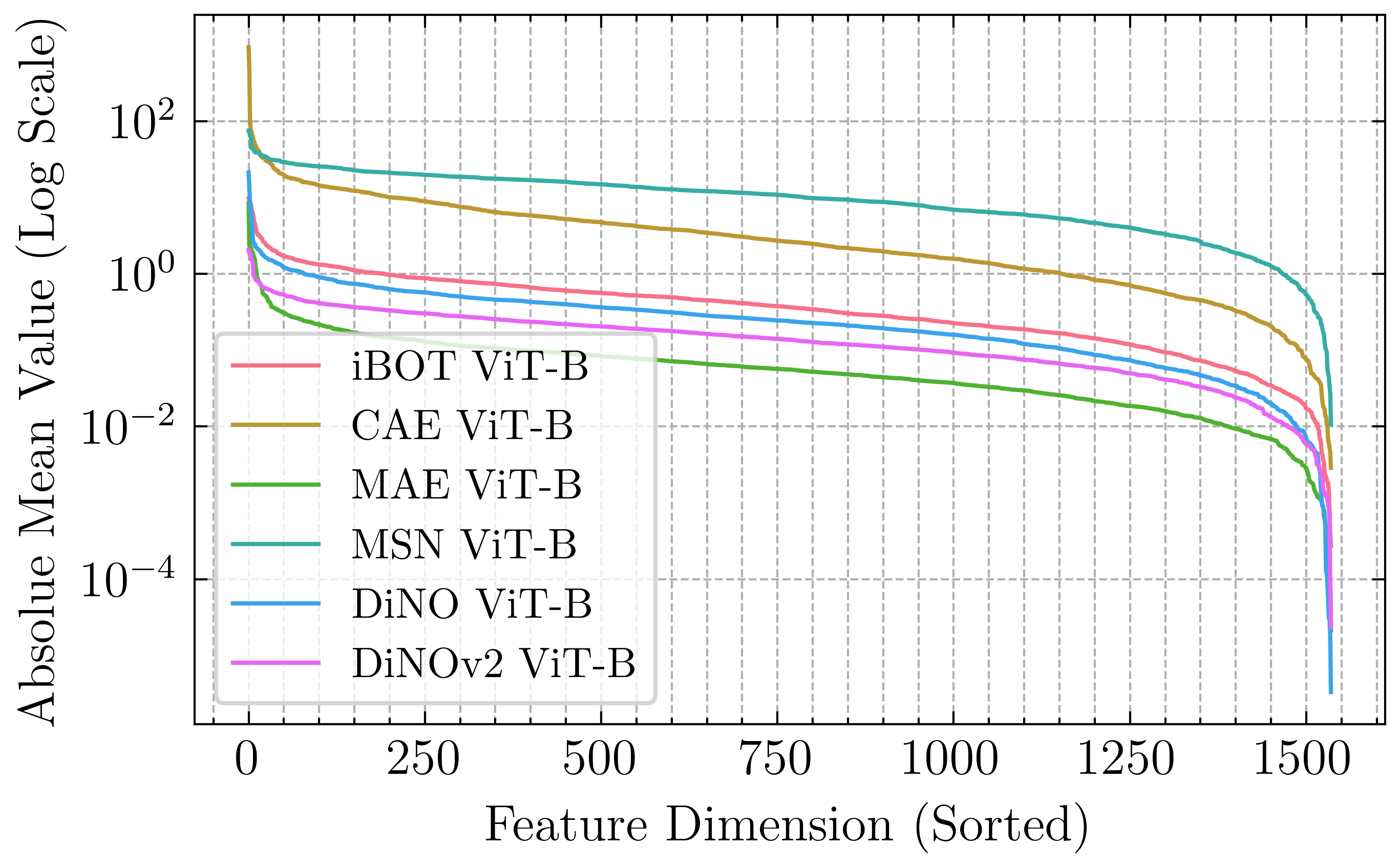}
        \caption{ViT-B models.}
    \end{subfigure}
    \begin{subfigure}{0.49\textwidth}
        \centering
        \includegraphics[width=\linewidth]{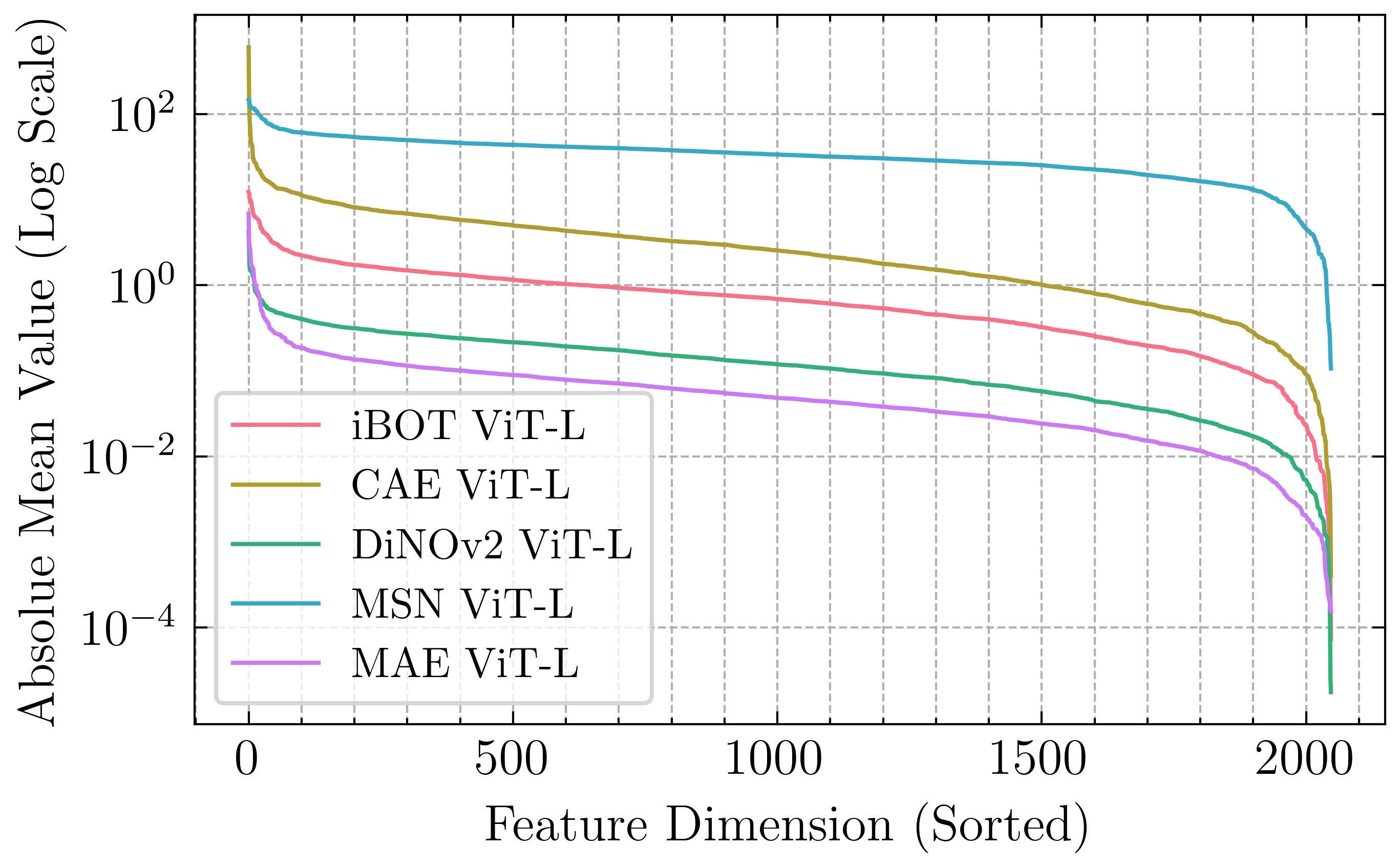}
        \caption{ViT-L models.}
    \end{subfigure}
    \caption{Absolute mean value per dimension for ViT-B (left) and ViT-L (right) VFMs. Feature coordinates are sorted in decreasing order of absolute mean value for features extracted from the ImageNet validation set.}
    \label{fig:vfms-means}
\end{figure*}

In \cref{fig:vfms-means}, we present the absolute mean value of each feature dimension for features extracted from the ImageNet validation set using ViT-B and ViT-L models. For each image, the feature vector is constructed by concatenating the class (\texttt{CLS}) token with the average patch token from the final layer. The results in \cref{fig:vfms-means} indicate significant variation in the absolute mean value across feature dimensions for all models. Additionally, for models such as DiNO ViT-B, CAE ViT-B, and MSN ViT-L, the mean feature vector is notably distant from the origin, underscoring the importance of mean-centering as a preprocessing step before computing the loss.

In \cref{tab:mean_centering}, we compare performances of TAAs against VFMs with and without the use of mean centering. Mean centering provides strictly better results compared to simply ignoring this processing step.

\begin{table}[htbp]
    \centering
    \caption{Classification accuracies on the Imagenette dataset for clean samples and TAA samples with and without the use of mean-centering.}
    \label{tab:mean_centering}
    \begin{tabular}{cccc}
        \toprule
         \textbf{Model} & \textbf{Clean} & \textbf{\textcolor{Orange}{No centering}} & \textbf{\textcolor{Orange}{Centering}} \\
         \midrule
         DiNO ViT-B & 85.0 & 5.6 & \textbf{5.4}\\
         DiNOv2 ViT-B & 90.5 & 0.4 & \textbf{0.2} \\
         MAE ViT-B & 71.5 & 9.2 & \textbf{6.2} \\
         MSN ViT-B & 85.8 & 10.4 & \textbf{9.9} \\
         \bottomrule
    \end{tabular}
\end{table}

\subsection{Impact of VFM layer selection for TAA} 
We demonstrate in \cref{tab:different_layer} and \cref{tab:different_layer_patch} that indeed, more efficient attack is created from the last layer of a VFM. It is not surprising, since the classification and segmentation heads are using the output tokens from the last layer as input. 
It is interesting to note that attacks carried against feature representations in middle layers perform the worst. We leave an explanation of this phenomenon for future research.

\begin{table}[htbp]
    \centering
    \caption{Classification accuracy, segmentation mIoU, and cosine similarity between original and adversarial \texttt{CLS} token for the last layer on the PascalVOC dataset for DiNOv2 ViT-S model when attacking \textbf{\texttt{CLS} token} with respect to layer from which it taken with PSNR equal to 40 db.}
    \label{tab:different_layer}
    \begin{tabular}{cccc}
        \toprule
         \textbf{Layer} & \textbf{Classification} & \textbf{Segmentation} & \textbf{\texttt{CLS} cos\_sim} \\
        \midrule
        
        \IT No attack & \IT 96.3 & \IT 81.4 & 1 \\
        \midrule

        1 (first) & 64.5 & 51.5 & 0.5 \\
        2 & 51.5 & 41.0 & 0.4 \\
        3 & 27.1 & 30.5 & 0.1 \\
        4 & 32.2 & 34.0 & 0.2 \\
        5 & 89.8 & 64.3 & 0.7 \\
        6 & 91.1 & 63.9 & 0.6 \\
        7 & 82.4 & 61.2 & 0.5 \\
        8 & 64.1 & 44.0 & 0.3 \\
        9 & 38.5 & 27.6 & 0.1 \\
        10 & 19.3 & 25.2 & 0.0 \\
        11 & \bf 6.2 & 22.1 & -0.0 \\
        12 (last) &  7.5 & \bf 19.1 & \bf -0.8 \\
        
        \bottomrule
    \end{tabular}
\end{table}

\begin{table}[htbp]
    \centering
    \caption{Classification accuracy, segmentation mIoU, and cosine similarity between original and adversarial \texttt{CLS} token for the last layer on the PascalVOC dataset for DiNOv2 ViT-S model when attacking \textbf{patch tokens} with respect to layer from which it taken with PSNR equal to 40 db.}
    \label{tab:different_layer_patch}
    \begin{tabular}{cccc}
        \toprule
         \textbf{Layer} & \textbf{Classification} & \textbf{Segmentation} & \textbf{\texttt{CLS} cos\_sim} \\
        \midrule
        
        \IT No attack & \IT 96.3 & \IT 81.4 & 1 \\
        \midrule

        1 (first) & 49.9 & 38.8 & 0.4 \\
        2 & 56.6 & 45.4 & 0.5 \\
        3 & 18.7 & 21.3 & 0.1 \\
        4 & 43.4 & 42.9 & 0.3 \\
        5 & 72.6 & 50.1 & 0.5 \\
        6 & 60.7 & 42.6 & 0.4 \\
        7 & 38.5 & 32.5 & 0.2 \\
        8 & 9.8 & 27.3 & 0.0 \\
        9 & 5.8 & 22.4 & 0.0 \\
        10 &6.3 & 23.1 & 0.0 \\
        11 & 2.0 & 14.1 & -0.0 \\
        12 (last) & \bf  0.1 & \bf 11.5 & \bf -0.3 \\
        
        \bottomrule
    \end{tabular}
\end{table}

\section{Extra experimental results}

\subsection{Transferability across models}
\begin{figure*}[ht]
    \centering
    \begin{subfigure}{0.49\textwidth}
        \centering
        \includegraphics[width=\linewidth]{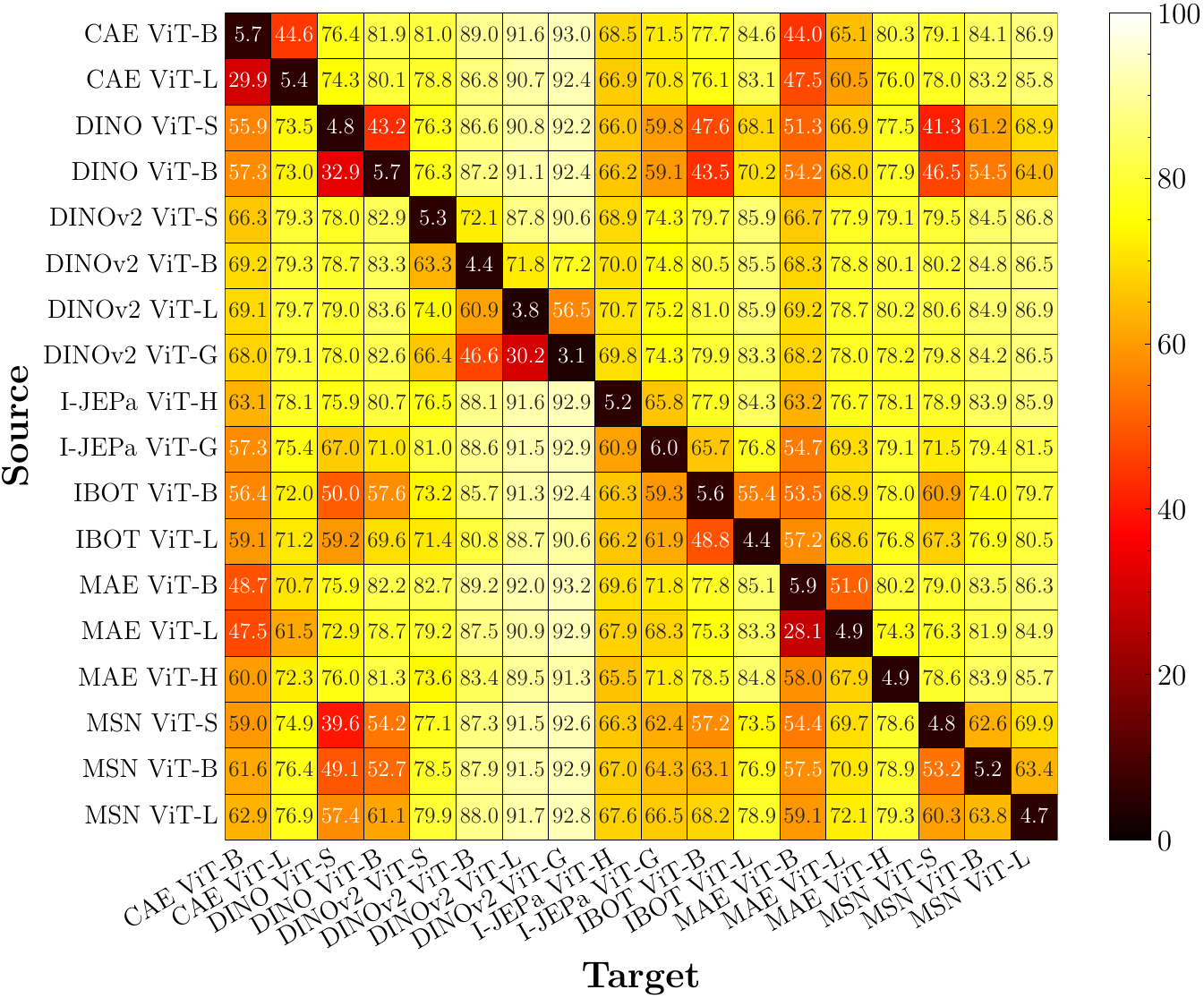}
        \caption{\textcolor{OliveGreen}{Task Specific Attacks (TSA)}}
    \end{subfigure}
    \begin{subfigure}{0.49\textwidth}
        \centering
        \includegraphics[width=\linewidth]{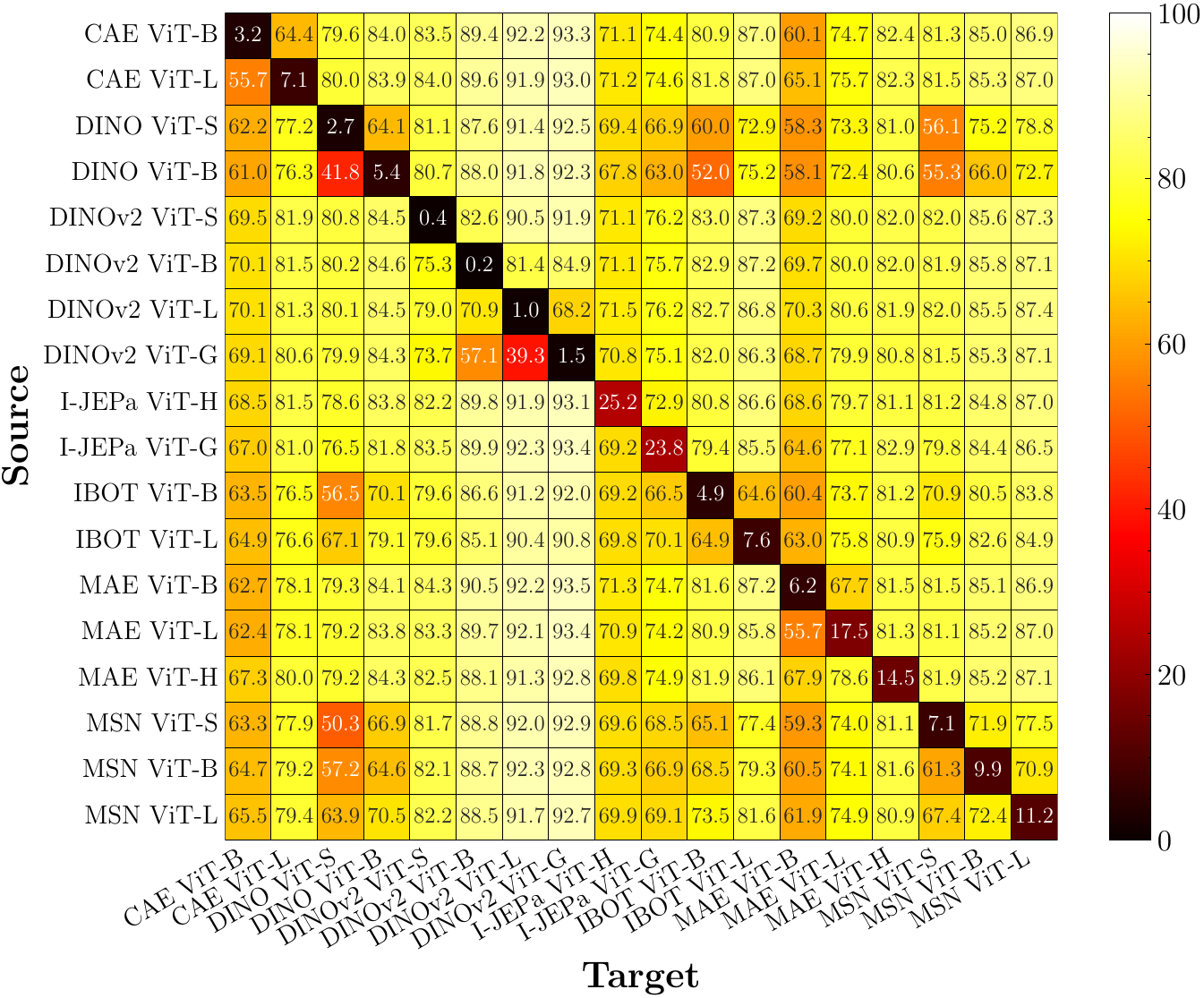}
        \caption{\textcolor{orange}{Task Agnostic Attacks (TAA)}}
    \end{subfigure}
    \caption{Absolute classification accuracy on the ImageNette dataset for TSAs (left) and TAAs (right)}
    \label{fig:models-table-class-abs}
\end{figure*}

\begin{figure*}[ht]
    \centering
    \begin{subfigure}{0.48\textwidth}
        \centering
        \includegraphics[width=\linewidth]{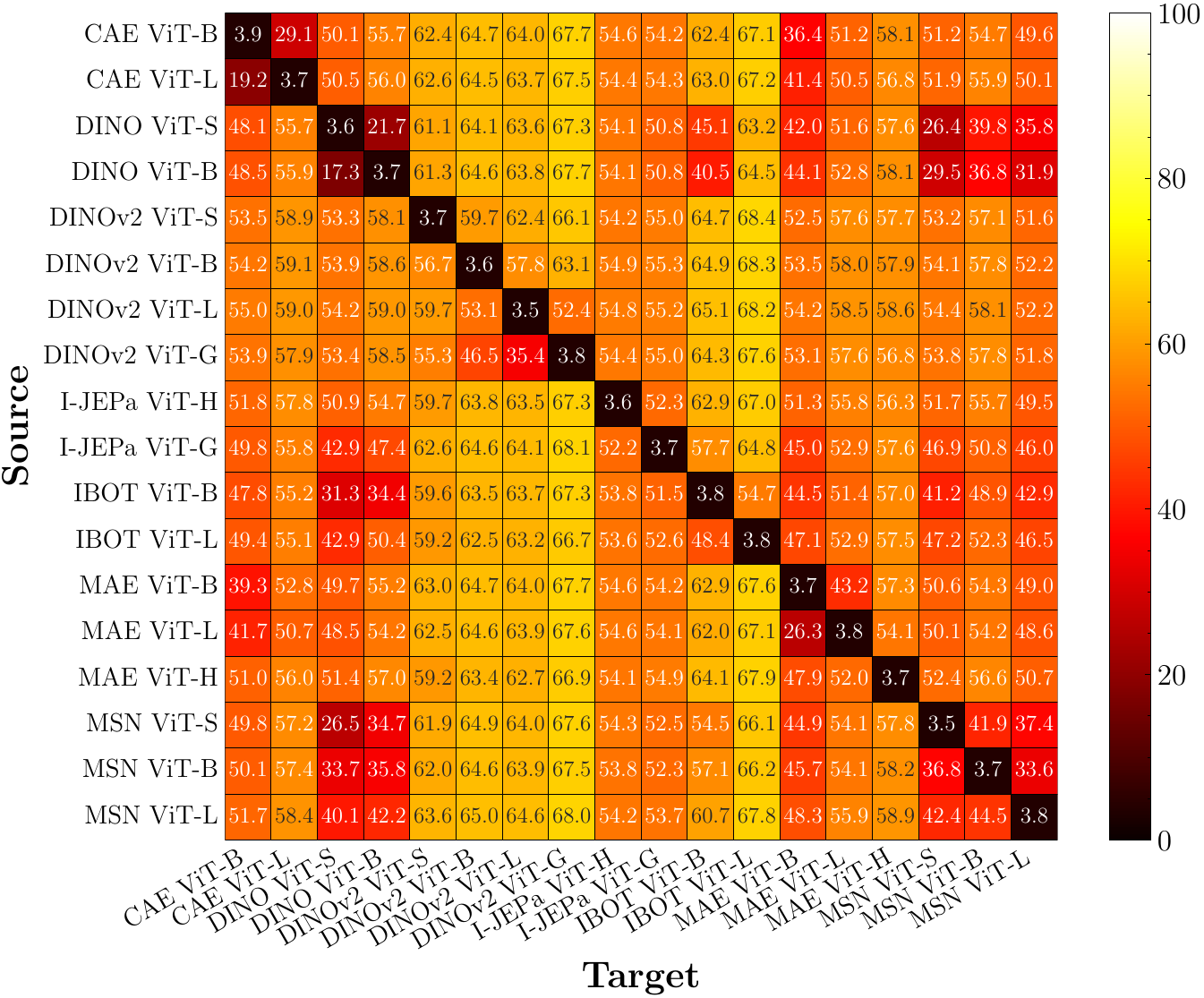}
        \caption{\textcolor{OliveGreen}{Task Specific Attacks (TSA)}}
    \end{subfigure}
    \begin{subfigure}{0.48\textwidth}
        \centering
        \includegraphics[width=\linewidth]{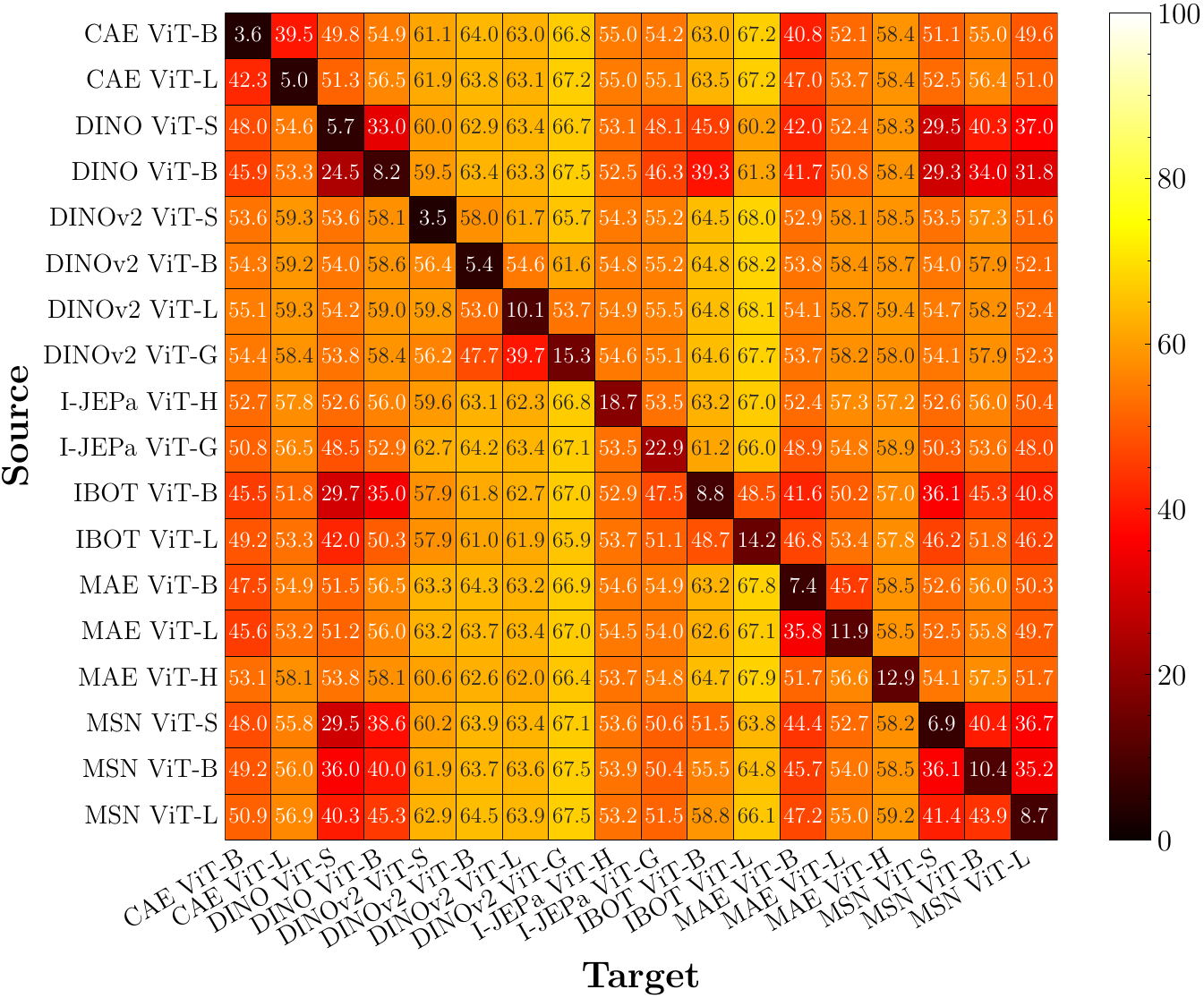}
        \caption{\textcolor{orange}{Task Agnostic Attacks (TAA)}}
    \end{subfigure}
    \caption{Absolute segmentation mIoU on the Pascal-VOC dataset for TSAs (left) and TAAs (right).}
    \label{fig:models-table-seg-abs}
\end{figure*}

We report the transferability of attacks across 18 models for Targeted Adversarial Attacks (TAAs) and Transferable Surrogate Attacks (TSAs) on both classification and segmentation tasks in \cref{fig:models-table-class-abs} and \cref{fig:models-table-seg-abs}, respectively. From these figures, we observe that while TAAs generally underperform compared to TSAs for their respective tasks, their performance remains comparable. Notably, for some model families, such as I-JEPA and MAE, TAAs significantly underperform relative to TSAs. However, in specific cases, such as classification with DiNOv2, TAAs demonstrate superior performance over TSAs.

\begin{figure}[htbp]
    \centering
    \includegraphics[width=\linewidth]{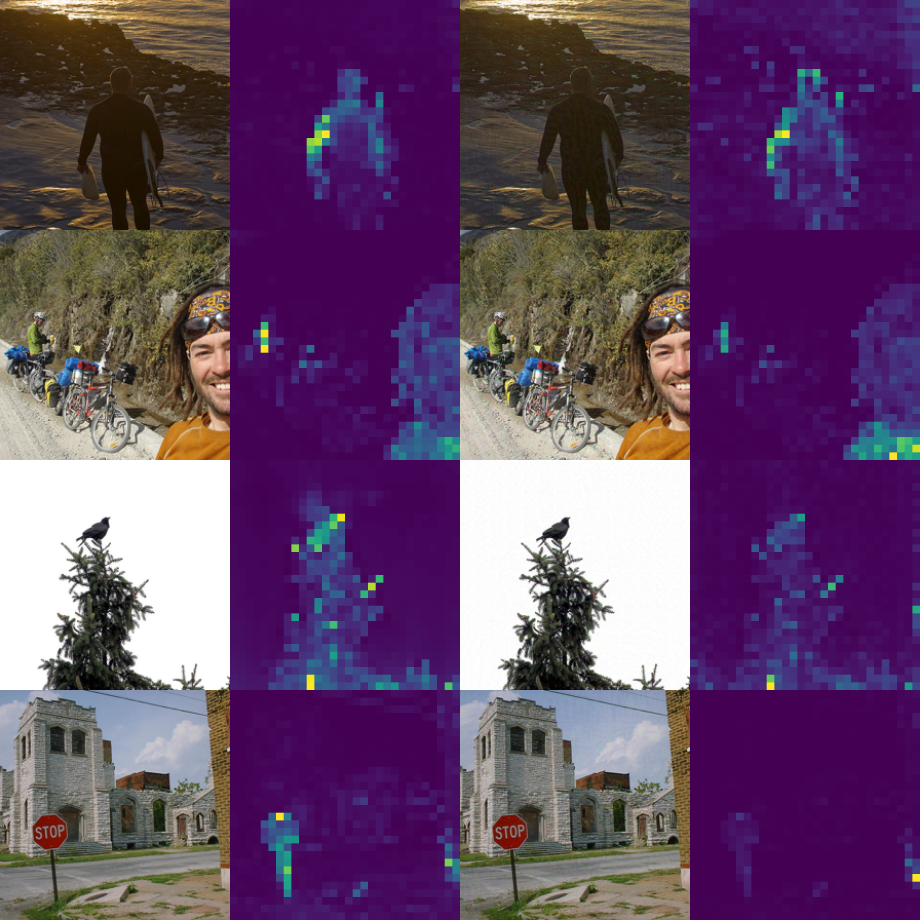}
    \caption{Corruption of attention masks of DiNO ViT-B using MSN ViT-B as a surrogate on the COCO2017 dataset. Original images and attention masks are in the first and second columns. Relative adversarial results are in the third and fourth columns. The target PSNR is 40dB.}
    \label{fig:att_masks}
\end{figure}

\subsection{Qualitative results for captioning and VQA}
\begin{figure*}
    \centering
    \includegraphics[width=\linewidth]{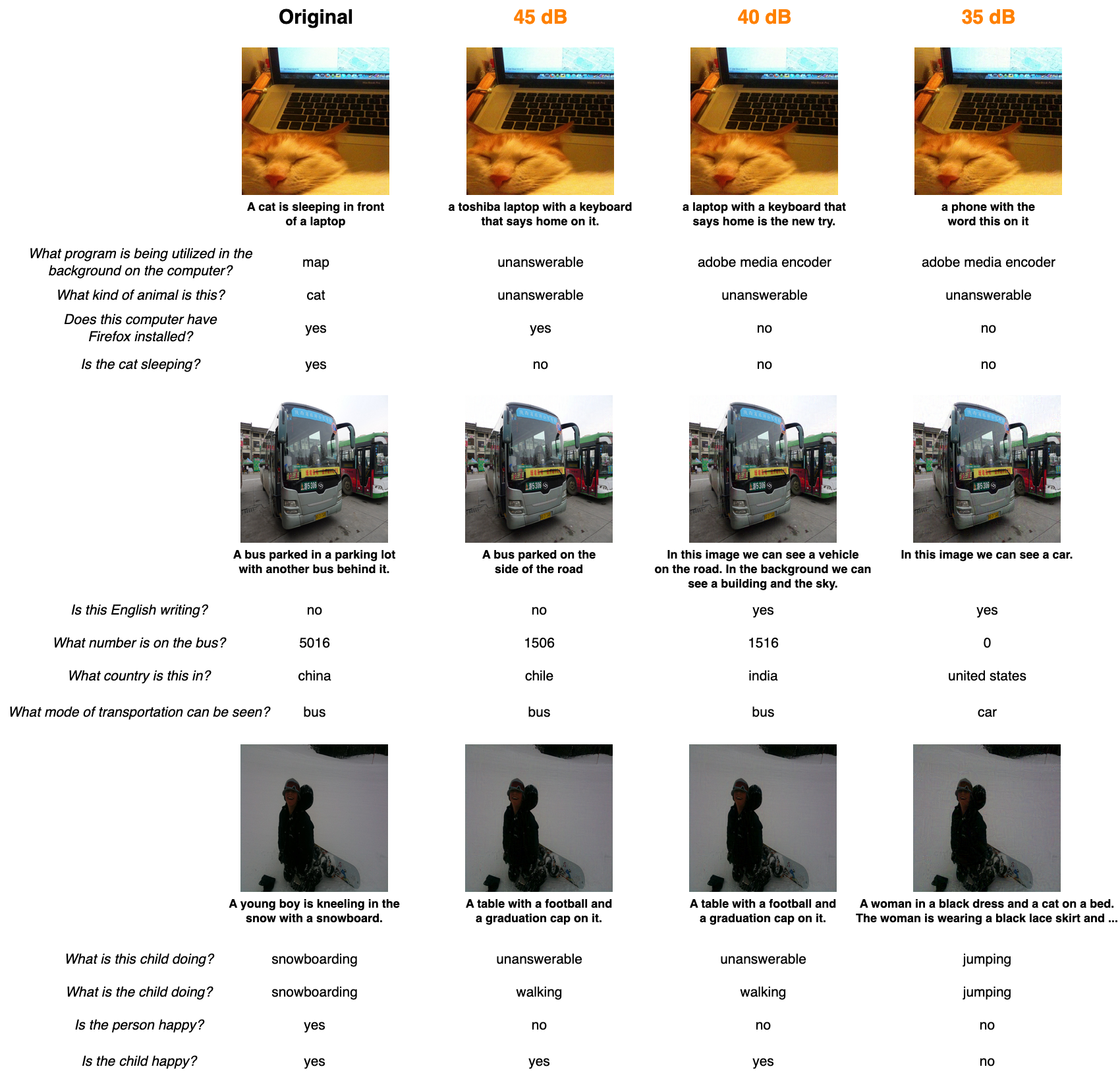}
    \caption{Example of captions and answers for images in the COCO dataset with TAAs using Paligemma.}
    \label{fig:vqa_cap_pali}
\end{figure*}

In \cref{fig:vqa_cap_pali}, we report a more exhaustive collection of captions and answers obtained with Paligemma for task-agnostic adversarial images on the COCO dataset.

\subsection{Qualitative results for classification and segmentation}
We present examples of outputs from the DiNOv2 ViT-S model, where linear layers were trained on top for classification and segmentation tasks, after applying various adversarial attacks targeting a PSNR of 40 dB, as shown in \cref{table:pred_all_attack}. While attacks based on the corresponding downstream loss yield the most effective results for the specific task, they exhibit poorer transferability to other downstream tasks. For instance, the third row illustrates an almost unchanged predicted segmentation mask when the attack is performed in the classification space, whereas an attack in the segmentation space fails to flip classification predictions in 3 out of 6 cases (rows 1, 2, and 5). Additionally, in classification attacks, the predicted segmentation contours remain nearly unaffected, with only the predicted class being altered (rows 1, 4, 5, and 6). Conversely, TAA %
effectively produce adversarial samples that significantly degrade model performance across multiple tasks.

\subsection{Analysing Self-attention} 
In \cref{fig:att_masks} we qualitatively visualize self-attention in DiNO on images from the MS-COCO2017 dataset. We show images before and after corruption in the black-box setting, using an MSN ViT-B model as the surrogate and a target PSNR of 40dB. Original images and attention masks are in the first and second columns respectively, while adversarial results are in the third and fourth columns. With this qualitative study, it emerges that slightly different images can indeed yield different self-attention maps with respect to the \texttt{[CLS]} token. We suspect that these manipulations can turn out to be harmful to segmentation and detection models based on features extracted with DiNO.

\subsection{Transferability across models}

We observe that the transferability of attacks across models is limited when we set the PSNR to 40dB. In \cref{tab:pgd-sia-mifgsm}, we study the transferability of TSA and TAA for classification under different adversarial attack strategies, namely PGD~\cite{DBLP:conf/iclr/MadryMSTV18}, SIA~\cite{wang2023sia} and MI-FGSM~\cite{dong2018mifgsm}, setting a lower PSNR value of 32dB.

The results in \cref{tab:pgd-sia-mifgsm} suggest that great improvements in transferability can be obtained with higher distortion levels and with advanced optimization algorithms. We leave a more thorough study of the transferability of TAAs across models for future work.

\begin{table}[ht]
\centering
\caption{Classification accuracy of the targeted models with attacks using model iBOT-B as a source model. PSNR$=32$dB. 
}
\label{tab:pgd-sia-mifgsm}
    \resizebox{0.48\textwidth}{!}{
    \begin{tabular}{*{8}{c}}
        \toprule
       Attack & iBOT-B & MSN-B &  MAE-B & DiNOv2-B \\
        \midrule
        PGD - TAA & \textbf{0.1} & 75.1 & 49.5 & 85.5 \\
        PGD - TSA & 5.5 & 63.3 & 44.3 & 83.1 \\
        MI-FGSM - TAA & 5.1 & 50.5 & 38.3 & 80.0 \\
        MI-FGSM - TSA & 5.5 & 42.8 & 33.1 & 69.4 \\
        SIA - TAA & 0.5 & \textbf{17.9} & \textbf{13.5} & \textbf{39.2} \\
        SIA - TSA & 2.5 & 20.7 & 17.6 & 42.1 \\
        \bottomrule
    \end{tabular}
    }
\end{table}

{
\clearpage
\onecolumn
\newcommand{\IMGSIZE}{0.195}

\newcommand{\image}[2]{\includegraphics[width=\IMGSIZE\textwidth,keepaspectratio]{Figures/examples/#1/#2.png}}

\newcommand{\imagesrow}[2]{\image{clean}{#1_clean_img} & \image{clean}{#1_gt_label} & \image{clean}{#1_pred_clean} & \image{patch_tokens}{#1_pred_adv} & \image{classification}{#1_pred_adv} & \image{segmentation}{#1_pred_adv}  \\}

\newcommand{\textrow}[4]{& #1 & #1 & #2 & #3 & #4 \\}

\begin{landscape}

\begin{longtable}{cccccc}
    \caption{Adversarial attacks for segmentation and classification downstream tasks for DiNOv2 ViT-S model with a target PSNR of 40 dB. \label{table:pred_all_attack}}\\
    \toprule
    
    \multirow{2}{*}{\textbf{Clean image}} & \multirow{2}{*}{\textbf{GT label}}        &
		\multirow{2}{*}{\textbf{Clean prediction}}       & \textbf{\TAA{TAA}} & \textbf{\TSA{TSA}} & \textbf{\TSA{TSA}} \\
    & & & \textbf{\TAA{patch tokens}} & \textbf{\TSA{classification}} & \textbf{\TSA{segmentation}} \\
		\midrule

    \imagesrow{3}{segmentation} 
    \textrow{bus}{bottle}{train}{bus}
    \midrule
    
    \imagesrow{7}{segmentation} 
    \textrow{bird}{dog}{aeroplane}{bird}
    \midrule
    
    \imagesrow{18}{segmentation} 
    \textrow{person}{bottle}{cat}{bottle}
    \midrule
    
    \imagesrow{26}{segmentation}
    \textrow{sofa}{motorbike}{chair}{tvmonitor}
    \midrule
    
    \imagesrow{27}{segmentation} 
    \textrow{horse}{cat}{cow}{horse}
    \midrule
    
    \imagesrow{29}{segmentation} 
    \textrow{train}{bottle}{bus}{person}
    
    \bottomrule

\end{longtable}

\end{landscape}

\clearpage
\twocolumn
}

\end{document}